# ComplAI: Theory of A Unified Framework for Multi-factor Assessment of Black-Box Supervised Machine Learning Models

Arkadipta De[*]
Indian Institute of Technology, Hyderabad, India
arkadipta.de@alumni.iith.ac.in

Satya Swaroop Gudipudi[*]
CVR College of Engineering, Hyderabad, India
gudipudiswaroop@gmail.com

Sourab Panchanan[*]
Pandit Deendayal Energy University, Gujarat, India
Panchanansourav@gmail.com

Maunendra Sankar Desarkar
Indian Institute of Technology, Hyderabad, India
maunendra@cse.iith.ac.in

## ABSTRACT

The advances in Artificial Intelligence are creating new opportunities to improve people's lives around the world, from business to healthcare, from lifestyle to education. For example, some systems profile the users using their demographic and behavioral characteristics to make certain domain-specific predictions. Often, such predictions impact the user's life directly or indirectly (e.g., loan disbursement, determining insurance coverage, shortlisting applications, etc.). As a result, the concerns over such AI-enabled systems are also increasing. To address these concerns, such systems are mandated to be responsible: transparent, fair, and explainable to developers and end-users. In this paper, we present ComplAI, a unique framework to enable, observe, analyze and quantify explainability, robustness, performance, fairness, and model's behavior in drift scenarios, and to provide a single *Trust Factor* that evaluates different supervised Machine Learning models not just from their ability to make correct predictions but from overall responsibility perspective. The framework helps users to (a) connect their models and enable explanations, (b) assess and visualize different aspects of the model such as robustness, drift susceptibility, and fairness, and (c) compare different models (from different model families or obtained through different hyperparameter settings) from an overall perspective thereby facilitating actionable recourse for improvement of the models. ComplAI is model agnostic and works with different supervised machine learning scenarios (i.e., Binary Classification, Multi-class Classification, and Regression) and frameworks (*viz.* scikit-learn, TensorFlow, etc.). It can be seamlessly integrated with any ML life-cycle framework. Thus, this already deployed framework aims to unify critical aspects of Responsible AI systems for regulating the development process of such real systems.

---

[*]Equal Contribution

## 1 INTRODUCTION

Machine learning algorithms are being used in almost every automated/assisted decision-making process in recent times. As many of these decisions significantly impact human lives and livelihood, the responsibility and transparency of these algorithms are becoming increasingly necessary. As a result, organizations are being mandated to regulate the development of compliant models concerning aspects of responsibility. Different countries, unions, and organizations have already started enforcing explainable, responsible, and trustworthy AI principles. The Department of Health and Human Services (HHS), U.S., has mandated the usage and deployment of responsible artificial intelligence systems in the healthcare domain[1]. World Health Organization (WHO) has published guidelines for responsible AI in the healthcare domain[2]. Similarly in India, National Institute of Transforming India (NITI)[3] and in Australia, Department of Industry, Innovation and Science have enforced principles of responsible AI. The European Union (EU) has established multiple individual, societal and systematic requirements that a trustworthy AI system must include.[4] The National Institute of Standards and Technology (NIST), U.S. Department of Commerce has mandated and defined four principles of Explainable AI[5] as follows:

- **Explanation**: Delivering accompanying evidence or reason(s) for all outputs.
- **Meaningful**: Providing explanations that are understandable to individual users.
- **Explanation Accuracy**: The explanation correctly reflects the system's process for generating the output.
- **Knowledge Limits**: The system only operates under conditions for which it was designed or when the system reaches sufficient confidence in its output.

Hence regulating these principles and making the machine learning models compliant with these guidelines has become a critical consideration for productization. Researchers are actively exploring approaches for enabling explainability, robustness, and fairness but the existing approaches function isolatedly. Moreover, within these solutions, some limitations may make them less feasible to use in the real world. Often organizations may not wish to divulge the internals of the models, yet they would like to maintain the regulations of these models. Hence, a unified framework to enable the aspects of responsible artificial intelligence has become essential.

---

[1]https://www.hhs.gov/sites/default/files/final-hhs-ai-strategy.pdf
[2]https://www.who.int/publications-detail-redirect/9789240029200
[3]https://www.niti.gov.in/sites/default/files/2021-02/Responsible-AI-22022021.pdf
[4]https://www.aepd.es/sites/default/files/2019-12/ai-ethics-guidelines.pdf
[5]https://www.nist.gov/system/files/documents/2020/08/17/NIST%20Explainable%20AI%20Draft%20NISTIR8312%20%281%29.pdf

In this paper, we propose a unified framework accompanying a graphical user interface suite that enables a large range of audiences to understand a model and the corresponding predictions. This framework will help data scientists develop machine learning models for compliant problems with the regulations of responsible artificial intelligence. It will allow the business analysts, QA engineers, and end-users to understand and explore the explanations on predictions of different data points.

The main contributions of this work are as follows:

- Extension of the NICE algorithm [3] for Counterfactual Explanations Generation to multi-class classification and regression problems.
- Providing counterfactual-based explanations for data points, class-wise robustness scores for classification and overall robustness scores for regression.
- Improving counterfactual flip-test for black-box fairness measurement to deal with small datasets in a better way.
- Introducing data drift analysis through counterfactuals to assess the model's susceptibility concerning Out-of-time (OOT) validation data.
- We compute an overall *AI trust score* for a model. It can be used to compare models for a particular problem.
- Finally, we present a graphical user interface for the framework that enables model explanations, performing assessments, and visualizing the machine learning model's performance from multiple aspects of responsible AI.

## 2 RELATED WORKS

Most of the current explainability methods are targeted toward specific model types (i.e., dependent on the model structure). Explainability for tree-based models based on game theory is presented in [11]. A linear classifier-specific method for generating actionable recourse is proposed in [17]. [15] proposed a method to efficiently find coherent counterfactual explanations using a novel set of constraints referred to as a "mixed polytope." LIME [14] was proposed to explain predictions of any classifier model in an "interpretable and faithful manner". A What-If Tool (WIT) was introduced in [21] that allows users to change feature values and explore the model outcome, explore fairness, and obtain the nearest real counterfactual from the dataset. These methods are black-box-centric, use-case specific, model type-specific, and very few are model agnostic. There have been some research works in the line of model robustness exploration and evaluation. [4] proposed new adversarial attack-based algorithms to measure the robustness of models. A novel robustness metric called CLEVER was proposed for neural network models in [20].

Counterfactual-based explainability methods have gained popularity and research attention. SEDC1 [12] was the first counterfactual evidence method and explained document classification. The method has been extended to behavioural data [13], images [18] and tabular data [7]. An autoencoder-based counterfactual generation method was introduced in [5]. The work in [10] minimizes prototype loss using autoencoder while generating counterfactuals, resulting in explanations closer to the data manifold.

All these methods enabling responsible AI work in isolation. The first attempt to bring all these aspects under one single framework was made in the CERTIFAI [16] framework. It uses counterfactual generation to arrive at explainability, robustness, and fairness scores through a genetic algorithm. However, the framework is used explicitly for classifiers, and the genetic algorithm for a counterfactual generation being computationally expensive makes the analysis heavily dependent on the small subset of data chosen for generating the scores. Also, [16] cannot perform well on high cardinal data as it genetically mutates the data.

Inspired by the foundations laid out by [16] and to address the shortcomings of the same, we propose **ComplAI** (pronounced as "comp-ly"), a counterfactual generation-based framework for assessing machine learning models. ComplAI can analyze explainability, robustness, performance, and fairness of a given model and scores the model on these aspects. It also allows data drift analysis and produces a single Trust Factor representing the overall model score. This enables different models to be compared from an overall perspective. The entire framework is efficient and scalable for large-scale real-world data. The framework is model agnostic and does not need to see inside the model to generate the scores. Table 1 compares **ComplAI** with existing works towards building a responsible artificial intelligence system.

## 3 COMPLAI FRAMEWORK

In this section, we describe the mathematical framework of ComplAI in detail. The framework is majorly based on the generation of counterfactuals. In real-world scenarios, the datasets are often large and can cater to various tasks such as binary classification, multi-class classification, regression, etc. We consider the NICE [3] algorithm for counterfactual generation and extend it further to (a) support and generate counterfactuals for large datasets and to (b) generate counterfactuals for multi-class classification and regression problems as well. At first, we introduce the NICE algorithm, its working methodology, and its extension to regression and multi-class classification problems. Then we introduce metrics enabling responsible AI framework in ComplAI.

### 3.1 Counterfactuals and Nearest Counterfactual

For a black-box model $f$, and a data-point $x$, a corresponding counterfactual is a feasible data-point $c$ of same dimension as $x$ such that the output of the black-box model for $x$ and $c$ are not equal. Extending this notion of counterfactuals, we define the nearest possible counterfactual of a datapoint $x$ for a black-box model $f$ to be a datapoint $c$ as,

$$\min_c d(x, c) \quad s.t. \ f(x) \neq f(c) \quad (1)$$

where $d(x, c)$ is the distance between $x$ and $c$. For example, if an algorithm assisting in loan grant decisions denies a loan for a particular candidate with a monthly income 5000\$ and we find a data point having all feature values the same as that of the previous data point, except that the monthly income is 7000\$, and the loan gets accepted by the system; then the later datapoint is a counterfactual of the former one.

| Method | Black Box | Model Agnostic | Mixed Data | Regression Support | Explainability/ Explainability Score | Robustness Score | Fairness Score | Drift Susceptibility Score | AI Trust Score |
|---|---|---|---|---|---|---|---|---|---|
| ComplAI | ✓ | ✓ | ✓ | ✓ | ✓ | ✓ | ✓ | ✓ | ✓ |
| CERTIFAI [16] | ✓ | ✓ | ✓ | ✓ | ✓ | ✓ | ✓ | | ✓ |
| Google What-If [21] | ✓ | ✓ | ✓ | ✓ | | ✓ | | | |
| Actionable Recourse [17] | ✓ | | ✓ | | ✓ | | ✓ | | |
| Counterfactual Explanations [19] | ✓ | ✓ | | ✓ | ✓ | | ✓ | | |
| Diverse Coherent Explanations [15] | ✓ | | ✓ | ✓ | ✓ | | ✓ | | |
| LIME [14] | ✓ | ✓ | ✓ | | ✓ | | | | |
| LORE [8] | ✓ | ✓ | ✓ | | ✓ | | | | |
| Adversarial Robustness [4] | | ✓ | | ✓ | | ✓ | | | |
| CLEVER [20] | | ✓ | | | | ✓ | | | |

Table 1: Comparing Related works with ComplAI. Mixed-data: Method works with both discrete and continuous data, without any discretization or assumptions, Regression Support: Support for Regression Model, Drift Susceptibility: Ability to detect drift in models and verdict whether model is susceptible to drift

Ideally, for a particular data point, a counterfactual is another data point just on the other side of a class's decision boundary, and the nearest counterfactual is the closest possible counterfactual to the chosen data point. Often in datasets used for building machine learning models, the *real* counterfactuals (i.e., another datapoint *from the dataset* having opposite outcome predicted by the model) available may not necessarily be the nearest counterfactual. Hence it is imperative to generate the nearest counterfactuals for better model assessment.

## 3.2 NICE Algorithm: Primer

NICE [3] is at the core of the ComplAI framework. It is a nearest counterfactual generation algorithm using the nearest neighbor-based approach. It generates synthetic counterfactuals from the actual instances in the training data.

Assume an $m$-dimensional feature space $X \subset \mathbb{R}^m$ consisting of both categorical and numerical features. A feature vector $x \in X$ has a corresponding class label denoted as $y \in \{0, 1\}$. A trained classification model $f : \mathbb{R}^m \to \{0, 1\}$ maps the input $x$ to a predicted output $\hat{y}_x$. A nearest counterfactual instance $c$ for $x$ minimizes a distance function $d(x, c)$ under the condition that $\hat{y}_x \neq \hat{y}_c$. The NICE algorithm uses Heterogeneous Euclidean Overlap Measurement (HEOM) [22] as this distance function ($d$) to minimize the distance between the actual data point ($x$) and counterfactual ($c$). The HEOM distance function for heterogeneous data-points (having both numerical and categorical values) for a specific feature $F$ is given by:

$$d_F(x, c) = \begin{cases} 1 & \text{if } x \neq c \text{ for Categorical Feature } F \\ 0 & \text{if } x = c \text{ for Categorical Feature } F \\ \frac{|x-c|}{\eta(F)} & \text{for Numerical Feature } F \end{cases} \quad (2)$$

Here $\eta$ is a normalization function for the numerical features. We use three different normalization methods: (a) range, (b) standard deviation, and (c) Mean Absolute deviation for numerical feature $F$.

For a datapoint $x$, the algorithm first selects the nearest neighbour $x_{nn}$ from the training set such that $\hat{y}_x \neq \hat{y}_{nn}$ and $y_{nn} = \hat{y}_{nn}$, which by definition is a counterfactual already. Next, the algorithm generates different synthesised datapoints $x_c$ over iterations using combinations of features values from $x$ and $x_{nn}$ to minimize the distance $d_F(x, x_c)$ between $x$ and $x_c$ and also ensure that $x_c$ remains a valid counterfactual of $x$ (i.e., $\hat{y}_c = y_{nn} \neq \hat{y}_x$). Additionally, over iterations, the algorithm also tries to maximize a set of reward functions $R(x_c)$ which enforces desirable properties of counterfactuals. Finally, the hybrid instance $x_c$ with simultaneously lowest distance value $d_F(x, x_c)$ and the highest reward $R(x_c)$ outcome is chosen as the generated synthetic nearest counterfactual $x_c$ of the datapoint $x$.

The reward functions $R(x)$ used along with the distance function $d_F(x, x_c)$ enforce certain desirable properties to the generated synthetic counterfactual $x_c$ such as Proximity, Plausibility, and Sparsity:

- **Proximity**: It indicates that the datapoint and the counterfactual should be close to each other, making the generated explanations easier to understand and take actionable recourse.
- **Sparsity**: It refers to the number of feature values that are different between the counterfactual and the actual data point. The less the number, the more sparse and convincing the explanation will be.
- **Plausibility**: It says that the values generated in the counterfactual must be plausible and not absurd. For example, in a loan application prediction, if a counterfactual says that "the experience must be changed from 2 years to 10 years to get the loan accepted", then it will not make much sense.

Detailed steps of the synthetic counterfactual generation algorithm along with the formulation of the reward functions used by NICE can be found in [3].

CERTIFAI uses genetic algorithms for generating nearest synthetic counterfactuals and hence is computationally expensive. Therefore, it may not work well for practical scenarios involving large datasets and high cardinality feature vectors. The NICE algorithm uses permutation and combination operations and is much light-weight. Also, it enables the generation of counterfactuals for large datasets more accurately as large datasets contain more variety of data.

## 3.3 Extension for Regression

We extend the algorithm to regression model where the output is a continuous variable. For a data-point $x_0$ and a black-box regression model $f$, we define the class of interest (i.e., Positive Class) to be in a range of $(f(x_0) - \lambda\sigma, f(x_0) + \lambda\sigma)$ where $\lambda$ is a user defined tolerance and $\sigma$ is the standard deviation of the target variable. Hence the nearest counterfactual $x_c$ for the data-point $x_0$ is defined

as:

$$\min_{x_c} d(x_0, x_c) \quad s.t \ f(x_c) \notin [f(x_0) - \lambda\sigma, f(x_0) + \lambda\sigma] \quad (3)$$

For example, consider the target variable as house price, and for a particular house data, it is 10000$. For $\lambda = 100$ and standard deviation $\sigma$ of the target as 10$, any house price below 9000$ or above 11000$ will be considered a counterfactual data-point. The upper and lower limits can also be customized using two different tolerance parameters (i.e., lower tolerance $\lambda_1$ and upper tolerance $\lambda_2$). Then the positive class (i.e class of interest) will have the range of $[f(x_0) - \lambda_1\sigma, f(x_0) + \lambda_2\sigma]$. In some cases, it may also be required that only all the higher or lower values be considered counterfactuals for an output. The formulation in Eq. (3) can then be modified accordingly to allow counterfactuals from the appropriate regions.

### 3.4 Extension for Multi-Class Classification

We also extend the NICE algorithm to multi-class classification problems. Let, a black-box multi-class classification model be $f$, and the set of classes be $C$. Then for a datapoint $x_0$, the corresponding counterfactual $x_c$ is defined as:

$$\min_{x_c} d(x_0, x_c) \quad s.t \ f(x_c) \neq f(x_0) \quad \text{where} \quad f(\cdot) \in C \quad (4)$$

For example, consider the case of plant species classification system $f$ where the objective is to classify the plant species out of 5 different species (say, $c_1, c_2, \ldots, c_5$) given data $x_0$ about a plant specimen. Also let the system classifies the specimen as $c_3$ (i.e., $f(x_0) = c_3$). Then, $c_3$ will be treated as positive class and the remaining classes viz. $c_1, c_2, c_4, c_5$ will be treated as negative classes. Hence, from (4) for datapoint $x_0$, the nearest counterfactual will be a datapoint $x_c$ such that the distance $d(x_0, x_c)$ between $x_0$ and $x_c$ is least and $f(x_c) \in \{c_1, c_2, c_4, c_5\}$.

### 3.5 Explainability

Explainability of an ML model refers to the concept of being able to understand the ML model. We define explainability by trying to answer the question "*What minimum change can be done to flip the prediction from/to a positive class (such as loan acceptance) to/from a negative class (such as loan rejection)?*" We consider the ability to answer these questions in the simplest and human interpretable form as *Explainability*. We enable it through nearest counterfactuals where, using the nearest counterfactual for a particular data point, the framework suggests what needs to be done to change the prediction. If it can be done by changing a few feature values, the explanation becomes simple. The definition of explainability and explanations easily extends to regression and multi-class classification techniques (as per Sec 3.3 and 3.4). By enabling counterfactual-based explainability, we can answer questions like "*Why was my loan application rejected?*" or "*What can I do to increase my insurance cover?*".

For this, we first generate the nearest counterfactual $c$ for every datapoint $x$ from the validation dataset $D_{val}$ and note the number of feature values ($n_f$) that need to be changed to arrive from $x$ to $c$. After that, we bucket these $n_f$ values into different bins (e.g., change in one feature, two features, etc.). We compute a weighted sum of the percentage occupancies of the bins, where the weight value decreases as we move to higher bins. This is based on the intuition that the most straightforward explanations need very few changes in the features. The simpler the explanation is, the easier it is to understand, and higher the score is awarded for the explainability of the model. The explainability score is calculated as:

$$Exp\_Score(f, D_{val}) = \sum_{\text{each bin } i} W_i P_i \quad (5)$$

where $W_i$ is the weight of the $i^{th}$ bin, and $P_i$ is the percentage of $n_f$ values falling in that bin. By default, we measure only up to 5 feature value changes, and the data points that need changes in more than five features are put into the final bin. This explainability score is then converted into a percentage, and a final explainability score is obtained. Additionally, we generate counterfactual based explanations of each instance that is given as an input to the framework.

| Feature | Original Value | Counterfactual Value |
|---------|----------------|----------------------|
| Age     | 50             | 52                   |
| BMI     | 25.3650        | 36.7650              |

Table 2: Example - Counterfactual based explanation in an Insurance Premium Prediction system (High or Low Premium). Explains when an individual needs to pay high premium. As indicated by explanation, he needs to pay 2 years later (i.e., when his age will be 52) and his BMI will also increase by 11.40.

Consider a loan application system $f$ with three input features viz. Annual Income, Years of Service, and Job Category (Business, Private, Government). For a particular person (i.e., datapoint) $x$, these feature values are 35000$, 6 Years and Private respectively. The system $f$ rejects the loan application of person $x$, and the person wants to know why the system rejected his loan application and what he should do so that his loan application is accepted. Through counterfactual analysis of ComplAI, we arrive at the nearest counterfactual[6] $c$ of $x$ for which the three feature values are 40000$, 7 Years and Private. Through this counterfactual analysis, the question asked by person $x$ can be directly answered, i.e., his annual income must be at least 40000$, and he must complete at least seven years of service to get his loan accepted. Thus, we enable explanations for any use-case (i.e., regression, binary classification, multi-class classification) through counterfactuals. Table 2 shows an example of such counterfactual based explanation generated through ComplAI.

### 3.6 Feature Importance

Feature importance helps us understand which feature(s) are dominating factors for decision-making for a particular problem. Besides showing the changes required to arrive at a favorable outcome through counterfactual explanations, we also compute the feature importance of the black-box model using generated counterfactuals of the validation dataset. We define the importance of a feature as the frequency of its change across counterfactuals of different datapoints. If a particular feature value changes in many counterfactuals, then we say that the particular feature is an essential and decisive factor for the model. For each datapoint $x$ and the corresponding generated counterfactual $c$, we observe which feature values are changing and keep track of the frequency of change for each feature across datapoints. The feature with the highest frequency of changing its values in counterfactual across all datapoints is the most crucial feature for that problem and the corresponding black-box model.

---
[6]which satisfies all rules of a favorable counterfactual, i.e., sparse, proximal to $x$ and plausible in nature

### 3.7 Transparency

Transparency of a Responsible AI system deals with the system's ability to explain the decisions made to the different stakeholders. This can be achieved by feature-based analysis for predictions and also by allowing the system user to change the input feature values to see whether the prediction changes.

We include slice-based analysis of model performance which enables the user to understand and debug the model performance in various parts (i.e., slices) of the data. ComplAI's UI empowers various stakeholders to query the model performance not only on the entire dataset but also on different data subsets by setting specific feature value ranges. This enables the identification of under-performing slices and also for slices for which there is not much training data.

The UI also comprises What-If analysis, which allows the user to edit any data-point and input custom values. Based on the custom values, the final model prediction, corresponding counterfactual explanation, and SHAP-based local feature importance, which features were positively and negatively responsible for the prediction, are displayed. This helps the system user to understand in detail the predictions by the models in a transparent way. Figure 1 shows how What-If analysis is performed for a particular problem.

### 3.8 Robustness

Robustness measures the ability of the algorithm to distinguish between data points from different classes. It measures how well a model has learned to identify instances from different classes and how hard it is to deceive the model. In terms of counterfactuals, robustness indicates how much effort is needed to change the outcome from one class to another class. To compute this score, for each point $(x_i, y_i) \in D_{val}$, we find the distance to its nearest counterfactual $c$ using distance function, $d_F(x, c)$ HEOM [22] as described in Section 3.2. For class $y_i$, we now define the robustness score as:

$$Rob\_Score_{y_i} = \mathbb{E}_{(x_i,y_i)}[d_F(x,c); x \in y_i, c \notin y_i] \quad (6)$$

These scores are then aggregated to obtain *minimum robustness score* over all classes and *average robustness scores* across all classes. The class-wise robustness score helps in understanding the model's robustness concerning a specific class. The class-specific score is helpful when the risk associated with misclassification with respect to a specific class is higher than others and also for class imbalance scenarios. In contrast, the aggregate value indicates the overall robustness of the model. A greater robustness score indicates a more prominent decision boundary learnt by the model.

### 3.9 Performance

The performance metrics used to evaluate any machine learning model depend on the task at hand (e.g., binary classification/multi-class classification/regression) and also on the distribution of the underlying data. There can be multiple evaluation metrics preferred by the data scientist for a single task. ComplAI allows the user to choose one or more metrics from an available large set of metrics, set importance to the chosen metrics and construct custom performance metrics *e.g.*, $0.3 * precision + 0.7 * F1$. We compute a single performance score based on these selections using the following equation:

$$Performance(f) = \sum_{p_i \in \mathbb{P}} w_i p_i \quad (7)$$

Where, $\mathbb{P} = \{p_1, p_2, \ldots, p_n\}$ is the set of performance metrics chosen by user and $w_i$s are the normalized importance weights of those performance metrics.

For regression problems, most of the scores are error based, i.e., Mean Squared Error, Mean Absolute Error, Mean Squared Log Error and follow the philosophy of lesser the score, better the model is. Hence for the regression problem, we base the performance score on either $R^2$ or Adjusted $R^2$ score as selected by the user.

### 3.10 Drift Detection through Counterfactuals

Usually, a machine learning model is trained on the training dataset and then tested on the testing dataset. In real-world scenarios, the deployed version of the trained model often observes data points that may not be coming from the observed data distribution. This "*drift*" may happen due to various reasons like environmental changes, effects of external or uncaptured contexts, etc. Most of the methods (in Section 2) suffer from Data and Label shifts [1]. If train data and Out-of-time (OOT) validation data/ live data are significantly different, there is no point in assessing a model's robustness, fairness, explainability, and performance. To understand the model's performance in these drift scenarios and the model's susceptibility to the drift scenarios, we use the concepts of **Data Drift**.

In **Data Drift** analysis, we try to understand how much the model's feature importance is changing according to the OOT validation data concerning the training data. We follow the method for data drift proposed by [9] where the feature importance attributes of the train data and OOT data are first calculated by our counterfactual method (see Section 3.6). Then a Normalized Discounted Cumulative Gain (NDCG) Score is calculated to measure the similarity in rankings between the feature importance of training data and OOT data. Let $F = [f_1, f_2, \ldots, f_n]$ and $F' = [f'_1, f'_2, \ldots, f'_n]$ be the counterfactual-based feature importance ranking on the training and OOT data respectively. $a(f)$ is a function that returns the feature attribution score for feature $f$. Then the corresponding NDCG, which we name *Feature Drift Sustainability Score* score is calculated as follows:

$$DCG = \sum_{i=1}^{n} \frac{a(f'_i)}{log_2(i+1)}, \quad iDCG = \sum_{i=1}^{n} \frac{a(f_i)}{log_2(i+1)} \quad (8)$$

$$Feat\_Drift\_Score(f, D_{val}, D_{live}) = \frac{DCG}{iDCG} \quad (9)$$

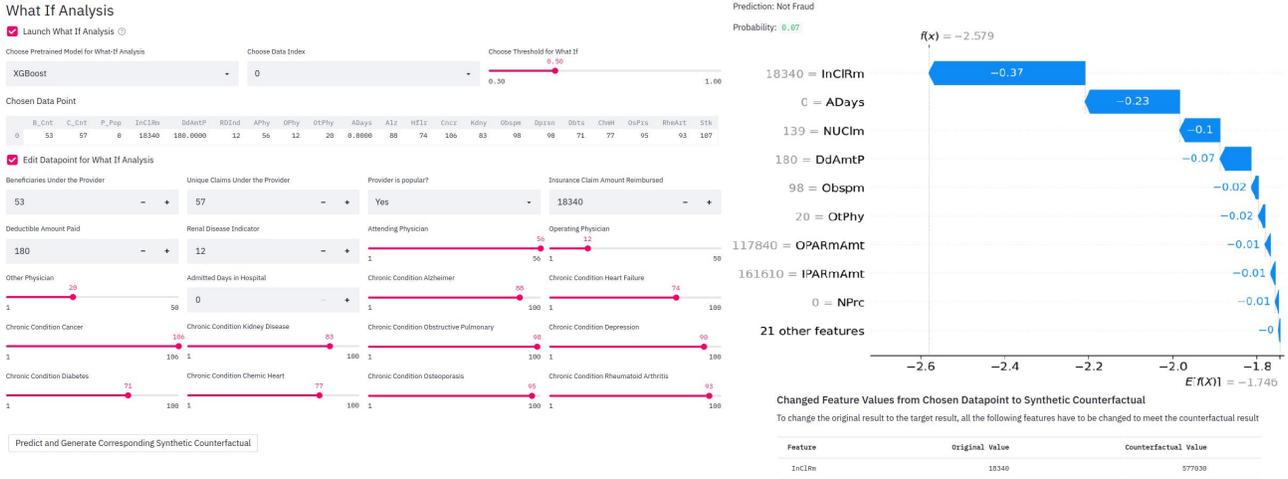

Figure 1: What-If Analysis in ComplAI framework

The *NDCG* score works as indicator of whether the model is performing well enough even for drifted data or it is susceptible to drift, moreover it helps to quantify the drift in the model. As NDCG measures the similarity between feature importance of training data and OOT data, higher NDCG score means less drift is detected in the model for the OOT dataset. If the *Feature Drift Score* is less than a threshold then we say a drift has occurred and trigger an alert. The OOT data can be cumulative or checkpoint based.

### 3.11 Fairness Analysis

Fairness is an important component of any responsible artificial intelligence system. For any black-box model, the model's fairness is measured over protected attributes (e.g. gender, sexual orientation, marital status, ethnicity, etc.). A well-established method for this task is the Counterfactual flip-test (FT) [2, 9] which assesses whether similar examples across the subgroups of a particular protected attribute receive similar predictions. For a user-defined protected attribute $f$, we partition the data into two groups $d$ and $a$ depending on the attribute values. For example, $d$ may contain the examples with *sex=female* and $a$ will contain examples with *sex≠female*. The members of facet $a$ are chosen to be the real/synthetic counterfactuals from the *k-nearest neighbors* of the observations from facet $d$. The possible outcomes are partitioned as *favorable* and *unfavorable* (e.g. class-1 vs class-0). Then the fairness quotient for the test is defined as:

$$FT = \frac{|F^+ - F^-|}{\text{Number of examples in } d} \quad (10)$$

where $F^+$ is the number of examples in $d$ with an unfavorable outcome whose nearest neighbors in $a$ have a favorable outcome. $F^-$ is defined analogously.

The counterfactual flip-test is entirely dependent on the quality of the available dataset as it finds and uses the nearest real counterfactuals within the dataset. Hence, it will not provide accurate results if the dataset is small. To overcome this, we generate synthetic counterfactuals while conducting Flip-Test (details in Sec 3.12). For a particular protected attribute and its value, generating synthetic counterfactuals that are not in the real dataset makes the facet $a$ much more populated and gives much closer counterfactuals than the real counterfactuals in $a$. Thus the final flip-test fairness score becomes more reliable and meaningful. We extend this flip-test framework for multi-class classification by choosing a class of interest as a favorable outcome, and all other classes are considered unfavorable. For regression, we consider a prediction range as the favorable outcome, and predictions outside this range are considered as unfavorable.

### 3.12 Counterfactual Flip-test Algorithm

Flip-test is a well studied algorithm for assessment of fairness of machine learning based classifier models. It tries to understand the question: *Will the model treat a datapoint differently if it had a different protected attribute value?* Here, we first define some useful definitions and then provide the pseudo code of our algorithm:

- **Favorable Outcome**: Class that is considered to be favorable according to the task at hand (usually denoted with class label 1). Example - Loan accepted, Heart disease not detected etc.
- **Unfavorable Outcome**: Class that is considered to be not favorable according to the task at hand (usually denoted with class label 0). Example - Loan rejected, Heart disease detected etc.
- **Privileged Group**: A set of datapoints having received a favorable outcome.
- **Unprivileged Group**: A set of datapoints having received an unfavorable outcome.
- **$F^+$**: For a particular protected attribute $p$, number of datapoints in the unprivileged group having a specific protected attribute value $v$ (i.e., $p = v$) whose nearest neighbour in the alternate group (set of datapoints having $p \neq v$) received a favorable outcome.
- **$F^-$**: For a particular protected attribute $p$, number of datapoints in the privileged group having specific protected attribute value $v$ (i.e., $p = v$) whose nearest neighbour in the alternate group (set of datapoints having $p \neq v$) received an unfavorable outcome.

The counterfactual flip-test algorithm in ComplAI evaluates fairness of a model from a particular protected attribute (viz. Gender, Race) perspective (i.e., it answers questions like: *How fair the model is with respect to protected attribute Gender?* In reference to Section 3.11, we describe our modified to the counterfactual based flip-test algorithm in detail below.

---

**Algorithm 1** Counterfactual Flip-test Algorithm for ComplAI

**Input:** $P$ = Set of protected attributes for a dataset $D$, $f$ = classifier trained on $D$
**Output:** Overall fairness score of classifier $f$ on dataset $D$ for the set of protected attributes $P$
1: **for** Each protected feature $p \in P$ **do**
2:    $X_{total}$ = Generate_Synthetic_Counterfactuals($p$) {Discussed in Algorithm 2}
3:    $X_{Norm}$ = Normalized $X_{total}$ {Normalized by feature-wise Mean Absolute Deviation}
4:    **for** Each feature value $v \in p$ **do**
5:      Create current subgroup, $S_i = \{x \mid x \in X_{Norm} \wedge x(p) = v\}$
6:      Create alternate subgroup, $O_i = \{x \mid x \in X_{Norm} \wedge x(p) \neq v\}$
7:      Initialize $k$-Nearest-Neighbour ($k$NN) on $O_i$ with $k = 1$
8:      $F^+ = |A|$ where, $A = \{(x, x_{nn}) \mid x_{nn} = kNN(x), x \in S_i, x_{nn} \in O_i, f(x) = 0, f(x_{nn}) = 1\}$
9:      $F^- = |B|$ where, $B = \{(x, x_{nn}) \mid x_{nn} = kNN(x), x \in S_i, x_{nn} \in O_i, f(x) = 1, f(x_{nn}) = 0\}$
10:      Flip-test value for $S_i$ is $FT_{S_i} = \frac{|F^+ - F^-|}{|S_i|}$
11:      Subgroup fairness score for $S_i$ is $F_{S_i} = (1 - FT_{S_i}) * 100$
12:    **end for**
13:    Fairness score for $f$ for $p \in P$, $F_p = min(F_{S_i}), \forall S_i$ of $p$
14: **end for**
15: **return** Overall fairness score for $f$ for $P$, $F_{Score} = min(F_p), \forall p \in P$

---

Here $f(x) = 1$ signifies favorable outcome and $f(x) = 0$ signifies unfavorable outcome for datapoint $x$. Next, we discuss the synthetic counterfactual generation algorithm which is used in the counterfactual flip-test algorithm 1. This generation algorithm is used to generate counterfactuals much closer to the original datapoints which helps in more accurate fairness assessment. It generates synthetic counterfactuals for all the subgroups of a protected attribute $p$ and finally returns a combined dataset containing all the original datapoints of the subgroups and corresponding generated synthetic counterfactuals that are not in the original dataset.

---

**Algorithm 2** Synthetic Counterfactual Generation for FlipTest

**Input:** $p$: A protected attribute, $D$: Dataset, $f$: the trained classifier
**Output:** $X$: Set with original datapoints and corresponding synthetic counterfactuals
1: Initialize $X = \{\}$
2: **for** Each distinct feature value $v$ of $p$ **do**
3:    Subgroup $S_v = \{x \mid x \in D \wedge x(p) = v\}$ {Subset of examples that have $p = v$}
4:    $S_v^{cf}$ = NICE($S_v$)
5:    Create synthesized subgroup $S_v^{syn} = \{x_{cf} \mid x_{cf} \in S_v^{cf} \wedge x_{cf}(p) = v\}$
   {Subset of $S_v^{cf}$ with $p = v$}
6:    $S_v^{final} = S_v \cup S_v^{syn}$ {Combines original and synthetic data}
7:    $X = X \cup S_v^{final}$
8: **end for**
9: **return** $X$ {Final combined data for protected feature $p$. Typically, $|X| >> |D|$}

---

For multi-class classification cases, we calculate fairness scores according to Algorithm 1 for each of the available classes as favorable outcome, keeping the other classes as unfavorable outcome, for each protected attribute $p$ respectively. The final fairness score is the minimum of fairness scores across all the protected attributes and across all the classes.

Counterfactual generation for subgroups can be optimized through parallel processing frameworks such as Dask[7], multiprocessing, etc and the overall complexity will be equivalent to KNN complexity with $k = 1$, i.e., $O(nd)$ where $n$ is the number of samples in the subgroup and $d$ is the dimension (i.e., number of features).

---
[7] https://www.dask.org/

### 3.13 AI Trust Factor

So far, we have described how the framework measures different aspects of the model. These scores are converted to percentages, and higher scores indicate better performance. The framework allows the users to select the set of scores to be considered and the importance weights of those scores to develop an aggregate score. This aggregate score, referred to as the **AI Trust factor (AI_Score)**, quantifies the model's overall performance on multiple aspects of a responsible machine learning algorithm, i.e., explainability, robustness, fairness, drift sustainability, and performance. It can be used to compare different models developed for the same problem definition. If $\mathbb{S} = \{s_1, s_2 \ldots, s_n\}$ is the set of chosen scores and $\mathbb{W} = \{w_1, w_2 \ldots, w_n\}$ is the set of corresponding weights then this score is computed as:

$$AI\_Score(\mathbb{S}, \mathbb{W}) = \sum_{i=1}^{n} w_i s_i \quad (11)$$

While equation (7) produces a unified score using weighted sum of different performance metrics of a model (viz. Accuracy, Precision, etc.), the equation (11) delivers a more complete assessment of the model from an overall perspective which is not restricted to performance only (as in equation (7)).

The weights selection ($\mathbb{W}$) for each of the model assessment factors varies with use case for example: a disease prediction model needs to be more explainable and fair and in contrast a churn prediction model can be reasonably explainable but highly performant. The weight selections of the model assessment factors for a model for a use-case entirely depends on the importance of these factors of the use-case and usually set by the subject matter (domain) experts of the organization.

## 4 INTEGRATION WITH ML LIFE-CYCLE AND PRODUCTION ENVIRONMENTS

Often organizations struggle to enforce Responsible AI (RAI) and try to document all meta-data used for model building before every production release. This approach cannot stop low quality and risky models from going to production due to absence of RAI as an automated quality check in CI/CD pipeline. To overcome this problem, we show how to integrate ComplAI with any Machine Learning life cycle framework seamlessly where a data scientist can build responsible models right from experimentation stage. The framework diagram is shown in Figure 2.

(1) The product owner/executive (PO) decides the business use case on which the ML solution will be built. The data scientist (DS) formulates the problem, makes a tentative list of required features for model creation, and submits these to the product owner. This is called model onboarding.
(2) Based on the model onboarding, the PO performs impact and risk assessment to decide the policy to be enforced while creating the models. The policy typically consists of threshold scores for Explainability, Robustness, Fairness, Drift, and Performance.

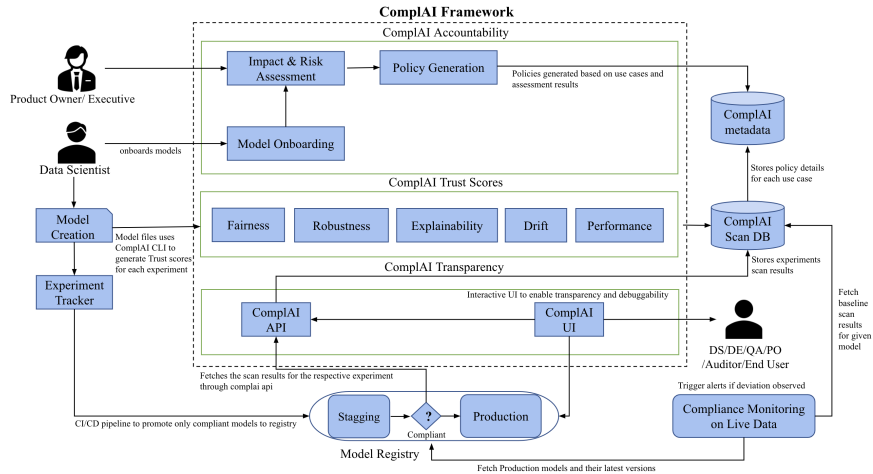

Figure 2: ComplAI Integrated with Machine Learning Life Cycle in Production Environment

These thresholds may vary depending on the use case. These policies are stored in ComplAI metadata storage.

(3) After policies are generated, data scientist(s) develop models. ComplAI Command Line Interface (CLI) can be used for each model to generate the trust scores using a single ComplAI command. These scores are then stored in the ComplAI Scan DB for each model along with the model artifacts (training data, validation data, hyperparameters, model files). All the models are tracked using the ML life cycle framework's (e.g. MLflow[23]) corresponding experiment tracker dashboard. The dashboard shows all the model scores (performance scores and all trust scores that ComplAI has generated).

(4) Finally, each of these models is sent through the Continuous Integration / Continuous Development (CI/CD) pipeline to the model staging environment of the ML life cycle framework. Each of these models is then automatically scanned through ComplAI API. A compliance check happens where each model's trust scores are compared with the policy (thresholds), and only compliant models are allowed to be sent into the production stage. The models sent to the staging phase can be accessed through the ComplAI User Interface. The UI displays different visualizations of the model statistics and allows users (PO, DS, Auditor, QA Engineer, and even End-User) to understand the model predictions, explanations, conduct what-if scenarios, etc.

(5) Even after the model has been sent to production, the ComplAI framework allows users to continuously monitor the model on live data and detect if feature drift occurs. Additionally, it allows users to continuously monitor if the model is compliant with respect to the ComplAI trust scores after deployment and detects any kind of trust score drift concerning the policy thresholds.

## 5 EXPERIMENTS ON OVERALL MODEL SCANS

For assessing the effectiveness of the ComplAI framework, we evaluate different ML models for different supervised tasks: Regression (Vehicle Pricing, Boston Housing Prices), binary classification (Lung Cancer Detection, Lead Scoring), and multi-class-classification (Iris Flower Species Detection, Wine Quality Classification). We use different ML models in this exercise: Linear Regression for regression problems, Logistic Regression for classification problems, and XGBoost, Support Vector Machine (SVR for regression and SVC for classification), and Multi-Layered-Perceptron for all the tasks. Explainability, robustness, performance, and overall AI Trust score are computed for each model. For calculating the AI trust score, we assign equal weight to each of the individual scoring components.

As the above datasets (except Lung Cancer Detection) do not have any protected attributes, we use different datasets to assess fairness (discussed in Section 6). For regression problems, we use adjusted $R^2$, and for classification problems, we use accuracy, F1 score etc. as performance metrics (varies with use-case). Ideally, users can use multiple different performance scores (accuracy, F1 score, recall, precision) and finally combine these into a final weighted average model performance score (discussed in Section 3.9). Table 3 shows the experimental results.

For the regression problems, XGBoost is most explainable for the Vehicle Pricing use-case, whereas MLP is most explainable for the House Pricing use-case. SVM and Linear Regression are most robust with **98.64** and **100** robustness scores in the Vehicle Pricing and House Pricing regression use-cases respectively. Also, Linear Regression and XGBoost are the best performant model obtaining the highest AI Trust Factor among the two regression problems respectively, which further enforces the fact that models must not be assessed based solely on one aspect, but multiple perspectives should be considered to decide the best performant. We see that SVM becomes the top-performing model from an overall perspective for Lead Scoring problem. SVM is generally a preferred algorithm for classification tasks, focusing on generalization through the max-margin principle. We also observe that, similar to the regression tasks, SVM has the highest robustness score for all the binary classification tasks since SVM aims to construct the optimal hyperplane that induces the largest margin. Also, since for all the models in binary classification use-cases, the performance score is very close to each other, the model quality differs in other aspects

| Problem Type | Dataset | Model | ComplAI Scores ||||||
|---|---|---|---|---|---|---|---|---|
| | | | Explainability Score | Robustness Score | Performance Score | Drift Susceptibility Score | Fairness Score | AI Trust Factor |
| Regression | Boston House Pricing | Lin Reg | 73.45 | 100 | 68.58 | 100.00 | NA | 85.51 |
| | | XGB | 70.05 | 94.46 | 88.63 | 97.00 | NA | 87.53 |
| | | SVR | 53.97 | 99.97 | 17.3 | 100.00 | NA | 67.81 |
| | | MLP | 79.71 | 94.52 | 50.33 | 98.00 | NA | 80.64 |
| | Vehicle Pricing | Lin Reg | 93.36 | 98.36 | 82.84 | 99 | NA | 93.39 |
| | | XGB | 99.43 | 47.25 | 92.48 | 99 | NA | 84.54 |
| | | SVR | 93.65 | 98.64 | 13.0 | 98 | NA | 76.32 |
| | | MLP | 94.34 | 86.66 | 37.49 | 99 | NA | 79.37 |
| Binary Classification | Lead Scoring | Log Reg | 89.6 | 44.23 | 91.21 | 100.00 | NA | 81.26 |
| | | XGB | 95.12 | 30.17 | 91.81 | 99.00 | NA | 79.03 |
| | | SVC | 89.33 | 47.03 | 90.16 | 99.00 | NA | 81.38 |
| | | MLP | 88.14 | 42.33 | 88.15 | 100.00 | NA | 79.63 |
| | Lung Cancer Detection | Log Reg | 84.29 | 95.92 | 75.64 | 84.75 | 97.01 | 87.52 |
| | | XGB | 74.68 | 89.79 | 87.18 | 90.00 | 99.00 | 88.13 |
| | | SVC | 56.35 | 100.00 | 87.18 | 82.61 | 98.00 | 84.83 |
| | | MLP | 81.35 | 73.56 | 82.05 | 95.88 | 99.00 | 86.37 |
| Multiclass Classification | Wine Quality Detection | Log Reg | 78.85 | 79.07 | 70.18 | NA | NA | 76.03 |
| | | XGB | 67.93 | 70.05 | 91.95 | NA | NA | 76.64 |
| | | SVC | 79.15 | 52.13 | 88.00 | NA | NA | 73.09 |
| | | MLP | 74.79 | 83.54 | 73.38 | NA | NA | 77.24 |
| | Iris Species Detection | Log Reg | 92.67 | 73.51 | 93.33 | NA | NA | 86.50 |
| | | XGB | 96.25 | 68.65 | 98.33 | NA | NA | 87.74 |
| | | SVC | 89.88 | 84.41 | 96.67 | NA | NA | 90.32 |
| | | MLP | 90.04 | 76.12 | 98.33 | NA | NA | 88.16 |

Table 3: Results of ComplAI Model Scan on different datasets. Performance Score varies for different use-cases (for regression adjusted $R^2$ and for all other cases F1 Score, Accuracy etc.); Since, the use-cases are from different domains (*viz.* Healthcare, Finance, Marketing etc.) we provide evidence of generalizability of the concept of ComplAI framework

such as explainability, robustness, etc. and hence ComplAI assessment is needed to understand which is the best model out of all the models for each of the use-cases. SVM achieves the highest AI Trust Factor for the Iris species classification dataset even though it is not highest performant model. This shows the importance of model assessment from an overall perspective rather than focusing on only performance scores.

| Model | ComplAI Scores ||||||
|---|---|---|---|---|---|---|
| | Explainability Score | Robustness Score | Drift Susceptibility Score (Mean) | Performance Score (F1 Score) | Fairness Score | AI Trust Factor |
| Log Reg | 97.67 | 3.20 | 97.9 | 92.48 | 80.77 | 74.45 |
| XGB | 98.49 | 58.26 | 87.74 | 93.40 | 90.31 | 85.69 |
| SVM | 94.72 | 88.42 | 85.23 | 61.94 | 93.37 | 84.68 |
| MLP | 93.65 | 89.46 | 98.65 | 95.30 | 63.06 | 88.28 |

Table 4: Results of ComplAI Model Scan on drift simulation scenario for Medical Insurance Dataset; Log Reg: Logistic Regression

To assess drift susceptibility scoring module of ComplAI, we choose Medical Insurance Dataset and introduce Gaussian noise in testing set to simulate drift scenarios. We maintain two versions of dataset, i.e., original version of training data using which all models were trained and drifted (noisy) version of testing data using which all models were tested. We assess our framework on this noisy dataset and report drift susceptibility scores along with all other scores. The result of this experiment is shown in Table 4. We observe that drift susceptibility is less for XGBoost and SVM. XGBoost is a boosting-based method, and SVM focuses on generalization - making the models less susceptible to drifts. On the other hand, Logistic regression and MLP have high drift susceptibility.

## 6 EXPERIMENTS ON FAIRNESS

We now compare our proposed fairness assessment metric against existing methods for fairness assessment, e.g., Disparate Impact (DI) [6] and Counterfactual Flip-Test (FT) [9]. Both these existing methods generate counterfactuals from the dataset itself and are heavily data dependent, especially for smaller datasets. We choose four different datasets - some with large number of examples and some with very less number of examples - for this analysis: Titanic Survival (890 samples), Adult Income (48000 samples), Medical Insurance (1338 samples), and Heart Disease Detection (303 samples).

| Data | CF Type | Fairness Method |||||||
|---|---|---|---|---|---|---|---|
| | | CF Flip-Test ||| Disparate Impact |||
| | | FT (Male) | FT (Female) | FT (Final) | DI (Male) | DI (Female) | DI (Final) |
| Titanic Survival | Existing | 0.34 | 0.29 | 0.29 | 0.08 | 0.82 | 0.16 |
| | Ours | 0.90 | 0.59 | 0.59 | 0.48 | 0.50 | 0.97 |
| Adult Income | Existing | 0.95 | 1.00 | 0.95 | 0.75 | 0.92 | 0.81 |
| | Ours | 0.98 | 0.96 | 0.96 | 0.54 | 0.56 | 0.95 |
| Medical Insurance | Existing | 0.93 | 0.97 | 0.93 | 0.29 | 0.20 | 0.69 |
| | Ours | 0.97 | 0.98 | 0.98 | 0.31 | 0.24 | 0.77 |
| Heart Disease Detection | Existing | 0.82 | 0.85 | 0.82 | 0.41 | 0.70 | 0.58 |
| | Ours | 0.98 | 0.83 | 0.83 | 0.50 | 0.48 | 0.96 |

Table 5: Comparison of fairness methods on protected attribute "Sex" using logistic regression. Final FT Score is the minimum of the FT scores over the facets (i.e. male and female); Final DI Score is the ratio of Min-DI and Max-DI over the facets.; Existing: Flip-Test (FT) and Disparate Impact (DI) with Real Counterfactuals; Ours: FT and DI with Synthetic Counterfactuals generated with algorithm described in Section 3.12

Logistic regression was used as the ML algorithm, and *sex* was considered to be the protected attribute. Table 5 shows the corresponding results. We see that the fairness score for the Titanic Survival dataset improves a lot when our method is used. The Flip-Test scores improves for all other datasets. In all the cases, we see a large increase in Disparate-Impact score as well. This is because real counterfactuals (used in existing methods) could be far from the original samples as the dataset is small, and they can be misleading in fairness comparison, as discussed earlier. Our method considers the closest possible synthetic counterfactuals in fairness computation and mitigates this data dependency. This impact is lesser for the Adult income dataset that has a larger number of samples.

## 7 ILLUSTRATIVE CASE STUDY

In this section, we present an illustrative case study of ComplAI. We present a simple case study with the Loan Approval Dataset that contains 983 labeled loan/credit applications. Features include applicant income, co-applicant income, loan amount, loan amount term, dependents, education, self-employed, credit history, property area type, gender, marriage and target variable loan status (Approved or Rejected).

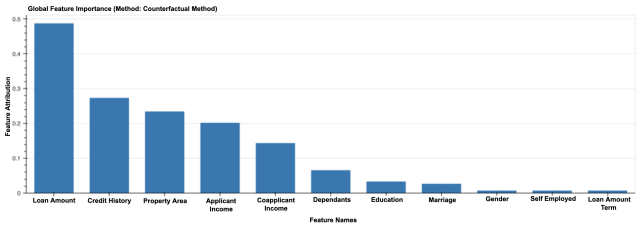

**Figure 3: Counterfactual based Feature Importance**

We train a logistic regression model on 615 samples as training data and analyze model behaviour on the remaining 368 samples. Since, the objective is to detect the applicants who can repay without defaulting hence false positive cases are to be heavily penalized and that is why we choose maximizing precision as the performance goal of the model. Figure 3 shows the ComplAI generated counterfactual-based feature importance where we see that credit history, loan amount and property area are the three most important features respectively.

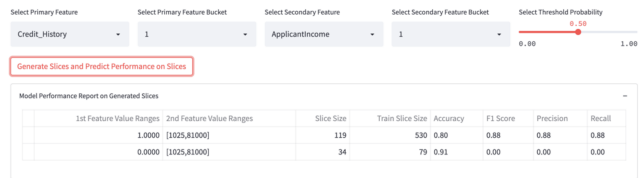

**Figure 4: Slice based Model Performance Analysis**

From overall model performance scan, we observe that the model achieves **88.46**% precision score which may appear to be satisfactory but using slice based analysis provided by the framework, we observe that the model is performing very poorly on the slice with credit history = 0 condition i.e., when no credit history is available for applicant (See Figure 4).

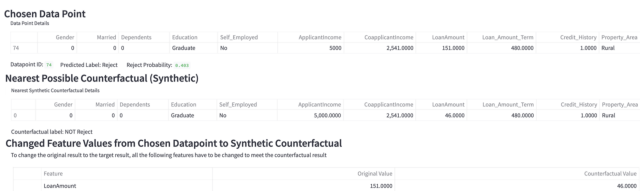

**Figure 5: Counterfactual based Explanation for a Data-point**

Generating explanations to questions such as *Why my loan application got rejected?* is very easy in ComplAI. In Figure 5, we choose an application where the original prediction is loan rejection and the framework generates the nearest synthetic counterfactual for the chosen data-point. We observe that the loan application was rejected because of the high loan amount. Had the loan amount been less then the application could have been accepted. Additionally, we also get an explainability score (See Section 3.5) of **92.86**%, average and minimum robustness score (See Section 3.8) of **98.35**% and **98**% respectively.

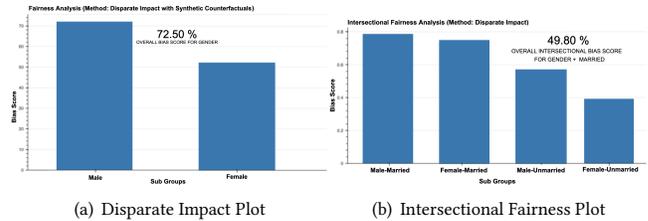

(a) Disparate Impact Plot  (b) Intersectional Fairness Plot

**Figure 6: Fairness Analysis in ComplAI on Protected Attributes**

ComplAI also allows fairness analysis of the model with respect to protected attributes. At first, we analyze Gender bias of the model through our framework. From Figure 6(a), we see there is a clear bias for Male subset of the dataset and model is under-performing for Female subset of the dataset. The disparate impact score for gender attribute is **72.50**%. The framework also enables in-depth fairness analysis with intersectional fairness analysis for multiple protected attributes. We analyze the intersectional fairness with two protected attributes *viz.* gender and marriage. From Figure 6(b), we see that the model is unfair for female subsets (as already observed from the gender bias analysis). Further, we observe that the model is especially unfair for female applicants who are unmarried. The intersectional fairness score is **49.80**%.

From the analysis of the model, we understand that even if performance metric precision is high, it is not enough to decide whether the model should be deployed on the use-case or not. Upon further analysis we understood model's incapability of performance for a specific slice. Moreover, we found out that the model is heavily biased towards the male subset of the data and under-performing for female applicants (especially for female applicants who are unmarried). Hence using different analysis modules of ComplAI, we arrive at the decision that the model cannot be deployed. Our proposed framework enables the capability of such detailed analysis of machine learning models, helps data scientists and business owners to arrive at crucial decision of model deployment and enables simple, interpretable explanations for the end user thereby inducing actionable recourse.

## 8 LIMITATIONS OF CURRENT VERSION OF COMPLAI

In this section, we discuss the limitations of ComplAI, some of which are being worked upon for inclusion in future versions of the framework.

- Currently, ComplAI generates a single nearest counterfactual for a given data point, but users may need multiple and diverse counterfactuals satisfying certain constraints.
- The current version of ComplAI works only with family of supervised machine learning models. It does not support unsupervised models such as Clustering algorithms.
- The framework currently supports labeled tabular datasets. It needs labels with the input data to generate counterfactuals. Currently it does not support unlabeled tabular data, textual dataset and image/video datasets.
- The framework does not support deep learning architectures such as CNNs, RNNs, LSTMs, Transformers, GANs etc. From the deep learning family of architectures, it only supports Multi-layered Perceptrons on labeled tabular datasets.

## 9 REPRODUCIBILITY

ComplAI is an already deployed software and it is being used in-house. It has been tested on multiple operating systems. In order to ensure experimental reproducibility of the results mentioned in the paper, we describe the requirements and steps necessary to run the prototype of ComplAI framework. We also provide details on how to reproduce each of the experimental results in Section 5 and 6 of the paper. Note that execution times may vary to a certain extent due to variations in individual hardware used to run the prototype.

### 9.1 Software and Hardware Specifications

A prototype version of ComplAI, along with all the necessary files for reproducing the experimental results from Section 5 and 6 is available in this *link* (use the Zip Password: Welcome2022!). A demo video on capabilities of the framework and integration with MLflow (with reference to Section 4) is also available in the link mentioned above.

The python libraries and package dependencies required to setup the prototype of ComplAI is defined in requirements.txt file of the prototype version of the framework.

For our individual setup and experiments, we used macOS Catalina 10.15.7 with 2.6 GHz 6-Core Intel Core i7 Processor, 16 GB of 2667 MHz DDR4 memory, Intel UHD Graphics Card 630 with 1536 MB memory, and 200GB of disk space.

### 9.2 Instructions to Reproduce Experiment Results

The following steps need to be executed to run the prototype of ComplAI and to reproduce the experiment results:

(1) Download the zip file from the *link* and unzip with the password: Welcome2022!
(2) To reproduce the experiment results in Table 3, go to the file complai_examples folder and follow instructions mentioned in README.md file.
(3) Complete setup to run ComplAI Scan (follow steps mentioned in "Steps to Run the code" section). Also complete setup for launching ComplAI web application (follow steps mentioned in "Steps to LAUNCH APP" section)
(4) After launching the web application in previous step, go to http://localhost:8501/ to access ComplAI's interactive UI (may also automatically redirect to the UI after launching the web app).
(5) To reproduce the experiment results in Table 5, go to the file fairness_experiments folder and follow instructions mentioned in README.md file.

## 10 CONCLUSION AND FUTURE WORKS

In this paper, we propose a unified framework for developing and enabling responsible artificial intelligence systems that are explainable, robust, fair, drift sustainable, and compliant to the organization and regulatory standards and can be seamlessly applied to use-cases of different domains (e.g., Healthcare, Finance, Marketing etc.). This framework is helpful for a wide range of audience i.e., enables data scientists to debug models at record level as well as at slice level from different perspective, enables quality assurance engineers to test ML models through What-If capability, helps product owners to review the model explanations from domain experts point of view, helps business specialists to understand what minimum effort is needed such that customer will buy a product (marketing use-cases) and finally enables end-users to contest and understand explanations such as - *Why my loan is rejected* etc. The current version of the framework allows different machine learning-based model families to be evaluated and can be integrated with ML life-cycle and production environments. In future, we aim to overcome the limitations of this framework and extend it to support textual data, image data, unsupervised models, and current state-of-the-art deep learning models such as RNNs, CNNs, LSTMs, and Transformers. In the future, we would also like to extend our framework to support the generation of diverse and constrained counterfactuals.